\documentclass[a4paper]{IEEEtran}
\usepackage{supertabular}
\usepackage{multirow}
\usepackage{booktabs, threeparttable}  
\usepackage{graphicx}
\usepackage{amssymb}
\usepackage{amsmath}
\usepackage{cite}
\usepackage{color}
\usepackage{epstopdf}
\usepackage{subfigure}
\usepackage{url} 
\usepackage{float}
\usepackage{makecell}
\usepackage{wasysym}
\usepackage[table,xcdraw]{xcolor}
\usepackage{diagbox}

\providecommand{\keywords}
\makeatletter
\newcommand{\figcaption}{\def\@captype{figure}\caption}
\newcommand{\tabcaption}{\def\@captype{table}\caption}

\makeatother
\hfuzz=\maxdimen
\begin{document}
\title{BASPRO: a balanced script producer for speech corpus collection based on the genetic algorithm}

\makeatletter 
\renewcommand{\section}{\@startsection{section}{1}{0mm}
  {-\baselineskip}{0.5\baselineskip}{\bf\leftline}}
\makeatother

\author{Yu-Wen Chen, Hsin-Min Wang, and Yu Tsao

\thanks{Yu-Wen Chen is with the Research Center for Information Technology Innovation, Academia Sinica, Taiwan and Department of Computer Science, Columbia University, New York, United States.}
\thanks{Yu Tsao is with the Research Center for Information Technology Innovation, Academia Sinica, Taiwan.}
\thanks{Hsin-Min Wang is with the Institute of Information Science, Academia Sinica, Taiwan.}
}

\maketitle

\textbf{\emph{Abstract} \---  
The performance of speech-processing models is heavily influenced by the speech corpus that is used for training and evaluation. In this study, we propose \emph{BAlanced Script PROducer} (BASPRO) system, which can automatically construct a phonetically balanced and rich set of Chinese sentences for collecting Mandarin Chinese speech data. First, we used pretrained natural language processing systems to extract ten-character candidate sentences from a large corpus of Chinese news texts. Then, we applied a genetic algorithm-based method to select 20 phonetically balanced sentence sets, each containing 20 sentences, from the candidate sentences. Using BASPRO, we obtained a recording script called \emph{TMNews}, which contains 400 ten-character sentences. \emph{TMNews} covers 84\% of the syllables used in the real world. Moreover, the syllable distribution has 0.96 cosine similarity to the real-world syllable distribution. We converted the script into a speech corpus using two text-to-speech systems. Using the designed speech corpus, we tested the performances of speech enhancement (SE) and automatic speech recognition (ASR), which are one of the most important regression- and classification-based speech processing tasks, respectively. The experimental results show that the SE and ASR models trained on the designed speech corpus outperform their counterparts trained on a randomly composed speech corpus.
}

\textbf{\emph{Index terms \--- corpus design, Mandarin Chinese speech corpus, phonetically balanced and rich corpus, recording script design, genetic algorithm.}}

\section{Introduction}
\IEEEPARstart{S}{}peech corpus plays a crucial role in the performance of speech-processing models. The speech corpus that is used to train and evaluate these models significantly affects their performance in real-world environments. Recently, massive amounts of data have been generated and collected. Therefore, models are often trained using a large amount of data to achieve better performance. However, not all research institutions can support such computing resources. Furthermore, the use of large amounts of data in listening tests to evaluate models is expensive and time consuming. Moreover, for personalizing models, the amount of data that can be collected from new users is often limited. Therefore, a representative speech corpus is essential for training and testing.

Active learning \cite{luo2021loss, bashar2021active} is a popular strategy used for training data sampling and selection. Active learning algorithm dynamically selects a subset of samples with labels that are most beneficial to improving the model during training. In this study, however, we focus on an algorithm that finds a fixed representative training and testing speech corpus for general speech-processing models. That is, active learning selects a corpus for a specific model to optimize it, whereas the proposed algorithm creates a model-independent corpus. The proposed algorithm can cooperate with active learning. Specifically, the model can be initially trained using the proposed representative corpus, followed by active learning to select the most beneficial samples for further training. 

A representative speech corpus is often referred to as a phonetically balanced or rich corpus. Phonetic balance means that the frequencies of phonemes in the corpus are distributed as close as possible to the frequencies in real-world conditions, and a phonetically rich corpus implies that the dataset should cover as many allowed phonemes as possible. In previous studies, researchers have developed corpora of this type for multiple languages, such as Amharic \cite{abate2005amharic}, Arabic \cite{abushariah2012phonetically}, Bangla \cite{ahmad2021sust}, Urdu \cite{raza2009design}, Thai \cite{wutiwiwatchai2002phonetically}, Turkish \cite{bozkurt2003text}, Mexican Spanish \cite{uraga2004voxmex}, Romanian \cite{stuanescu2012asr} and Chinese \cite{wang1998statistical,liang2003efficient, zhang2008design}.

Previously, phonetically balanced and rich corpora were designed by experts with linguistic backgrounds \cite{kurematsu1990atr,zue1990speech, huang2005development}. The experts manually wrote or chose sentences that could form a phonetically balanced corpus. However, creating a phonetically balanced and rich corpus in this manner is time-consuming and difficult. In addition, sentences written by the same person tend to be similar and lack variation. Moreover, this method cannot be used to generate corpora for specific knowledge domains.

Automatic methods have also been proposed, in addition to manual development. Automatic methods usually begin with a large collection of sentences. An algorithm then selects sentences from the collection to form a corpus that meets these requirements. Selecting the desired set of sentences is an NP-hard set-covering optimization problem. In other words, evaluating all possible sets of sentences is computationally too complex to be solved within an acceptable time. To automatically compose a phonetically balanced corpus, \cite{oh2011phonetically,stuanescu2012asr} proposed random sampling and evaluating sentence groups and chose the one that best meets the requirements. \cite{abate2005amharic} and \cite{wang1998statistical} proposed two-stage methods. The first stage selects important sentences that contain as many syllables as possible or consist of units that appear less frequently in the corpus. The second stage involves selecting sentences that can achieve the desired statistical distribution. Additionally, \cite{villasenor2004experiments} used the perplexity of each sentence as an indicator to generate a corpus. Most automatic methods are based on greedy algorithms \cite{bozkurt2003text, liang2003efficient,zhang2008design,raza2009design,ahmad2021sust}. Genetic algorithms (GA), a well-known approach for solving NP-hard problems, on the other hand, have not received much attention in speech corpus development. 

In \cite{tsai2009development}, the authors proposed a GA-based method to automatically form a phonetically balanced Chinese word list; nevertheless, this study focused on word lists rather than sentence lists. Only a few previous studies have used GA to automatically select sentence sets \cite{nicodem2007recording, nicodem2008evolutionary}. Moreover, these GA-based methods focus on phonetic and prosodic enrichment rather than phonetic balance and enrichment. The development of GA-based Chinese speech corpora has not yet been thoroughly investigated.
 
Mandarin Chinese is a tonal syllabic language with five different tones, including four main tones and a neutral tone. Syllables that do not consider tone are denoted as base syllables. On the other hand, syllables that consider the tonal information are referred to as tonal syllables. Each syllable comprises an INITIAL (consonant) and a FINAL (vowel) and is represented by the pinyin system. The INITIAL and FINAL can be further decomposed into smaller acoustic units such as phonemes. Compared to phonemes, syllables are more intuitive to Mandarin Chinese speakers and are used more frequently. Therefore, we developed a tonal syllable-balanced and -rich (hereafter referred to as syllable-balanced) corpus to represent a phonetically balanced and rich corpus.

In this study, we propose an automatic method called BAlanced Script PROducer (BASPRO)\footnote{The toolkit is available via: \url{https://github.com/yuwchen/BASPRO}} to compose a syllable-balanced Mandarin Chinese speech corpus. First, BASPRO uses pretrained natural language processing (NLP) systems to extract candidate sentences from a huge Chinese news text corpus. Subsequently, a syllable-balanced recording script is generated using a GA-based method. Finally, the script is converted into a speech corpus using two text-to-speech (TTS) systems. The syllable-balanced recording script developed in this study is called \emph{TMNews}\footnote{The script is available via: \url{https://github.com/yuwchen/BASPRO/tree/main/TMNews}} because the sentences in the script are collected from Mandarin Chinese news articles collected in Taiwan. 

The contributions of this study are as follows.
\begin{itemize}
\item We propose BASPRO, which uses machine-learning-based NLP tools to process and extract candidate sentences from a collection of news articles. 
\item BASPRO employs a GA-based method to form a syllable-balanced recording script from candidate sentences. Experimental results show that the proposed BASPRO system can effectively select sentences according to the designed optimization criteria. 
\item The proposed BASPRO system is flexible in terms of language, data domain, and script size. In addition, it allows the generated script to have multiple sets, each satisfying the desired requirements. For example, in this work, each of the 20 sets is syllable-balanced, and the sentences do not overlap between sets. 
\item We analyze the performance of speech processing models trained on syllable-balanced (produced by BASPRO) and randomly composed speech corpora. Experimental results show that the speech processing models trained on the syllable-balanced corpus perform better than those trained on the randomly composed corpus.
\end{itemize}

\section{The Proposed BASPRO System}
\begin{figure}[htbp!]
\centerline{\includegraphics[scale=1.]{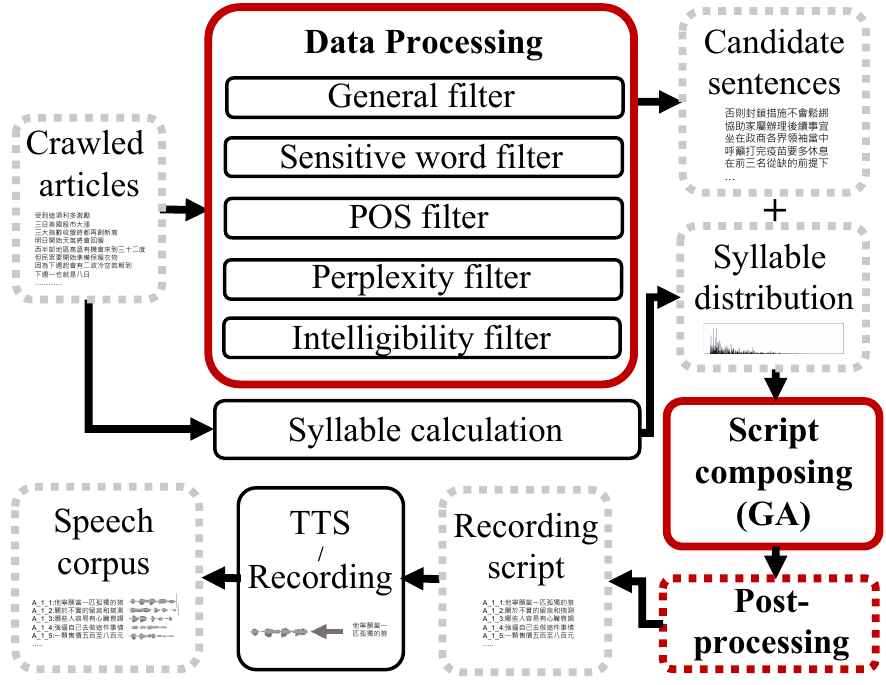}}
\caption{Schematic diagram of the proposed BASPRO system. In the data processing phase, candidate sentences were extracted from the collected articles. The script-composing phase uses real-world syllable distribution to compose a syllable-balanced script from the candidate sentences. Finally, the postprocessing phase replaces unwanted sentences in the script and produces the final script.}
\label{fig:proposed_method}
\end{figure}

The proposed BASPRO system consists of three main phases: data processing, script-composing, and postprocessing. The input is articles crawled from the Internet, and the output is a syllable-balanced recording script. Speech corpora can be generated from recording scripts using TTS systems or by asking people to make recordings. Figure~\ref{fig:proposed_method} shows a schematic of the BASPRO system. First, the data processing phase extracts candidate sentences from the collected news articles. Simultaneously, the syllable distribution of the collected articles was calculated, which is denoted as real-world syllable distribution. The script-composing phase then generates a temporary syllable-balanced script from the candidate sentences. Finally, the postprocessing phase replaces unwanted sentences in the temporary script and produces the final script.

\subsection{Data processing}
In the data processing phase, the input is news articles crawled from the Internet, and the output is candidate sentences. All sentences in the recording script were selected from the candidate sentences. We used five filters in the data processing phase to extract candidate sentences: (1) general, (2) sensitive word, (3) part-of-speech (POS), (4) perplexity, and (5) intelligibility filters. The general filter removes sentences with non-Chinese characters and keeps sentences with exactly ten characters. The sensitive filter then removes the sentences containing sensitive words. In this study, we let the sentences have a fixed length and excluded sentences containing sensitive words, as these settings are often required for listening tests. In addition, we designed a POS filter, perplexity filter, and intelligibility filter to filter out incomprehensible sentences. Because the resulting corpus will be used for listening tasks, we do not want any sentences to be difficult to understand and thus affect the evaluation results.

The POS is a category of lexical items with similar grammatical properties. Words assigned to the same POS often play similar roles in the grammatical structure of a sentence. We used POS as an indicator to exclude sentences that may not be suitable for listening tests. For example, a sentence containing a proper noun may be difficult to understand for someone who has never heard the word before, leading to a personal bias in listening tests. Meanwhile, sentences that start with a preposition, particle, or conjunction, and sentences that end with a preposition or conjunction are also inappropriate because they are usually not complete sentences. Therefore, we used two pretrained POS tagging systems to tag candidate sentences and remove sentences that met the above POS-based removal criteria.

Perplexity (PPL) is defined as the model’s uncertainty regarding a sentence. Higher perplexity indicates that a sentence may be more difficult to understand. In this study, we used pre-trained BERT\cite{devlin-etal-2019-bert}, a neural-network-based model trained with a masked language modeling objective, to compute the perplexity of each sentence. Given a sentence $W = (w_1, . . .,w_i, ..., w_{|W|})$, $w_i$ is the $i$-th character in $W$. To calculate $W$’s perplexity, $w_i $ is replaced with the [MASK] token and predicted using all other characters in $W$, that is, $W_{ \setminus i}= (w_1,..., w_{i-1}, w_{i+1},..., w_{|W|})$. $P_{BERT}(w_{i}|W_{\setminus i})$ is the probability of $w_i$ given its context calculated by BERT. Then, the perplexity of sentence $W$ is defined as:

\begin{equation}
PPL(W)=-\frac{1}{|W|}\sum_{i=1}^{|W|}\log P_{BERT}(w_{i}|W_{\setminus i})
\end{equation}

A high $PPL(W)$ indicates that $W$ contains characters that are difficult to predict from their context, suggesting that $W$ can be difficult to understand. We computed the perplexity for each sentence and analyzed the distribution of perplexity across all sentences to determine a threshold. The perplexity filter then removes sentences whose perplexity is above the threshold.

\begin{figure}[htbp!]
\centerline{\includegraphics[scale=1.]{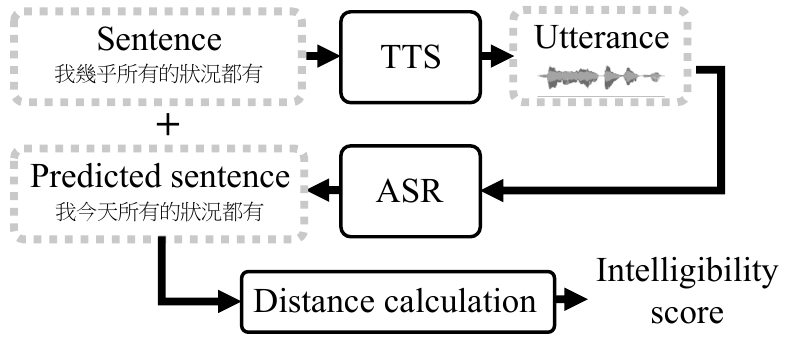}}
\caption{Illustration of the intelligibility score calculation. First, a sentence is converted into an utterance using a TTS system. Then, an ASR system is used to predict the content of the utterance. The distance between the sentence and ASR prediction is used to calculate the intelligibility score.}
\label{fig:intel_filter}
\end{figure}

The last is the intelligibility filter, which removes sentences with low intelligibility scores. Figure~\ref{fig:intel_filter} illustrates the calculation of the intelligibility score for a sentence. First, a TTS system was used to convert a sentence into a corresponding speech utterance. Subsequently, a pretrained automatic speech recognition (ASR) system is used to predict the content of the utterance. Finally, the Levenshtein distance between the sentence and the ASR prediction is used to measure the intelligibility of the sentence. If a sentence is difficult to understand, the TTS system may not be able to generate a correctly pronounced utterance because some characters have multiple pronunciations. In addition, previous research \cite{chen2021inqss} showed that ASR predictions are highly correlated with human perception of intelligibility. In other words, if a sentence is confusing, the ASR system may fail to correctly recognize the corresponding speech utterance. Therefore, the distance between ASR prediction and the original sentence reflects the intelligibility of the sentence. The intelligibility score is defined as one minus the distance of the sentence divided by the length of the sentence. Therefore, a perfect ASR prediction will lead to an intelligibility score of 1.

\subsection{Script-composing}

In the script-composing phase, we used the GA to select sentences, from the candidate sentences, to form a syllable-balanced recording script. The script consisted of several sets, each containing a fixed number of sentences, and the sentences did not overlap between sets. First, we introduce the basic concept of the GA. Then, we present the proposed GA-based script-composing method. 

\subsubsection{Genetic algorithm (GA)}
The GA is inspired by natural selection—a process of eliminating the weak and leaving only the strong. In the GA, the population is a series of possible solutions named chromosomes. Chromosomes are composed of genes that represent specific items. A fitness function is used to evaluate each chromosome. The fitness score reflects how well a chromosome ``fits'' the problem; a higher fitness score indicates that the chromosome is a better solution.

The GA comprises five steps: (1) initialization, (2) fitness calculation, (3) selection, (4) crossover, and (5) mutation. The initialization step creates the initial population and the fitness calculation step calculates the fitness score of each chromosome in the population. In the selection step, chromosomes with higher fitness scores have higher probabilities of leaving their offspring in the next generation. In the crossover step, a pair of selected chromosomes exchanges genes to form a new pair of chromosomes. Take one-point crossover as an example, a point called ``crossover point'' on both parents' chromosomes is randomly chosen. Then, the genes to the right of the crossover point are swapped between the parent chromosomes, producing two new chromosomes that carry genetic information from both parents. Lastly, genes in chromosomes may change randomly during the mutation step.

\subsubsection{The GA-based script-composing phase}

\begin{figure}[htbp!]
\centerline{\includegraphics[scale=1.]{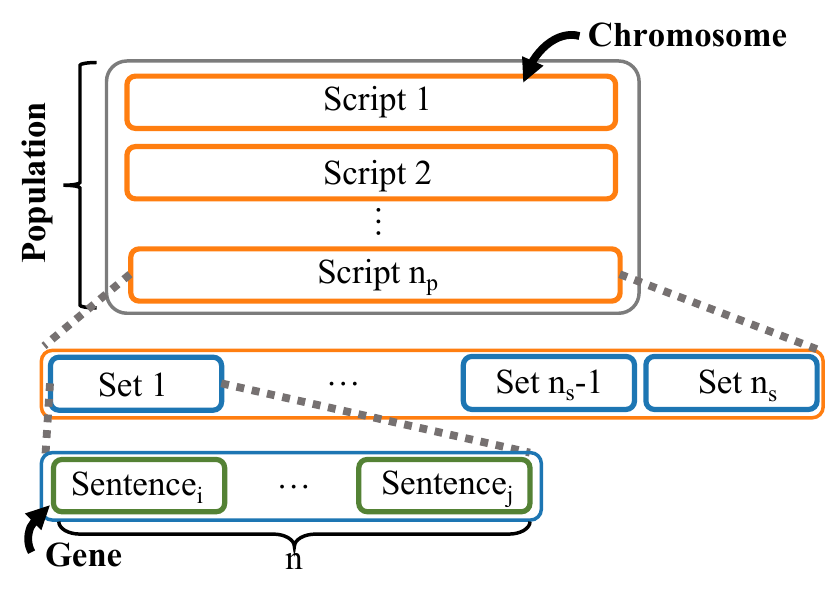}}
\caption{GA terms and their corresponding definitions. The \emph{population} is a collection of scripts, each \emph{chromosome} is a script, each \emph{gene} is a sentence, and $n_p$, $n_s$, and $n$ denote the number of scripts in the population, number of sets in a script, and number of sentences in a set, respectively. Sentence$_i$ denotes the $i$-th sentences in the candidate sentence set. The sentences were randomly sampled from the candidate sentence set during initialization, and there were no duplicate sentences in each script.}
\label{fig:ga_definition}
\end{figure}

Figure~\ref{fig:ga_definition} shows the GA terms and the corresponding definitions in this study. The \emph{population} comprises a collection of scripts. Each \emph{chromosome} is a script and the best chromosome in the population is the target syllable-balanced script. A \emph{gene} is a sentence that is swapped between chromosomes.

\begin{figure}[htbp!]
\centerline{\includegraphics[scale=.9]{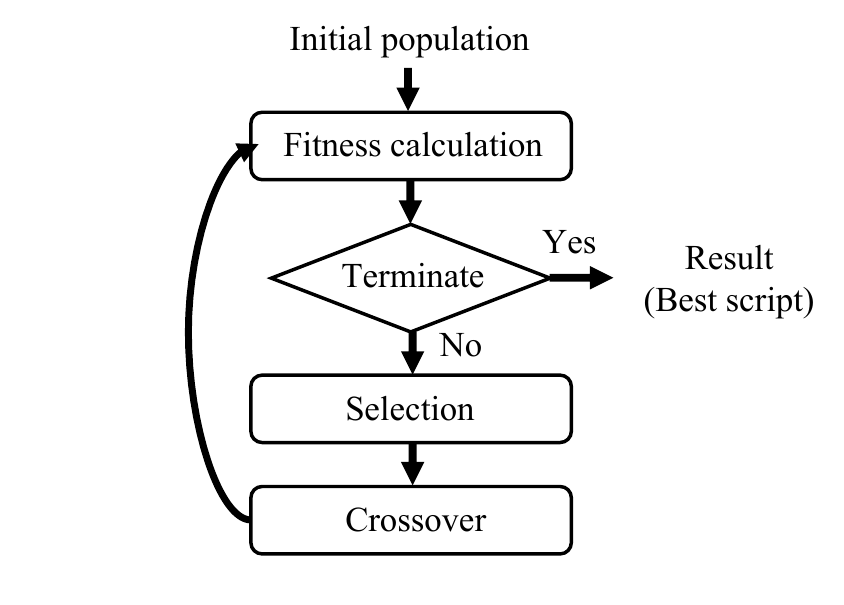}}
\caption{Schematic diagram of GA. The termination condition occurs when the maximum fitness score no longer increases after several generations.}
\label{fig:ga_steps}
\end{figure}

Figure~\ref{fig:ga_steps} illustrates the GA process. The initial population step generated multiple scripts, each consisting of random sentences. The fitness calculation step then calculates the fitness score of each script in the population. The selection step replaces scripts with lower scores with scripts with higher fitness scores. The crossover step exchanges sentences between the scripts. This process stops when the population is dominated by one script and the maximum fitness score no longer increases. We skip the mutation step because it increases the complexity without improving the performance of our test.

\subsubsection{Fitness calculation}	
The fitness calculation step evaluates how well a script satisfies the requirements. Specifically, a script with a higher fitness score is considered a better choice. In this study, the fitness score is defined as follows:

\begin{equation}
\begin{aligned}
Fitness\_score =  w_1\times script\_syllable\_distribution \\ +  w_2\times script\_syllable\_coverage \\ +  w_3\times set\_syllable\_distribution
\end{aligned}
\label{eq:fitness}
\end{equation}
where $w_1$, $w_2$, and $w_3$ are the weights.

Let $D_{script}$ be the syllable distribution of a script and $D_{real}$ be the real-world syllable distribution, $D_{script}$ $\in$ $R^s$, $D_{real}$ $\in$ $R^s$, and $s$ be the number of distinct syllables in Mandarin Chinese. The script\_syllable\_distribution is the cosine similarity between $D_{script}$ and $D_{real}$.

\begin{equation}
script\_syllable\_distribution=\frac{D_{script} \cdot D_{real}}{\left \| D_{script} \right \|\left \| D_{real} \right \|}
\end{equation}

Similarly, the set\_syllable\_distribution is the average cosine similarity between the real-world syllable distribution and each set in the script.

\begin{equation}
set\_syllable\_distribution=\frac{1}{n_s}\sum_{i=1}^{n_{s}}\frac{D_{set}^{i} \cdot D_{real}}{\left \| D_{set}^{i} \right \|\left \| D_{real} \right \|}
\end{equation}

where $D_{set}^{i}$ is the syllable distribution of the $i$-th set in the script, and $n_s$ is the number of sets in the script. We include the set\_syllable\_distribution in the fitness score such that each set is representative and can be used individually. For example, each set can be used as a validation set in the training of a speech-processing model and as an indicator for selecting the best model. Additionally, each set can be used for model training when only a small amount of data is required.

Script\_syllable\_coverage is the fraction of all possible syllables covered in a script. For example, assuming that the number of distinct syllables in Mandarin Chinese is 1300, the script\_syllable\_coverage score of a script that contains 130 distinct syllables is 0.1 (i.e., 130/1300). Note that in this study, we consider \emph{tonal} syllables instead of base syllables. In other words, the fitness function calculates the distribution and coverage of the tonal syllables.

\subsubsection{Selection}
\begin{figure}[htbp!]
\centerline{\includegraphics[scale=1.]{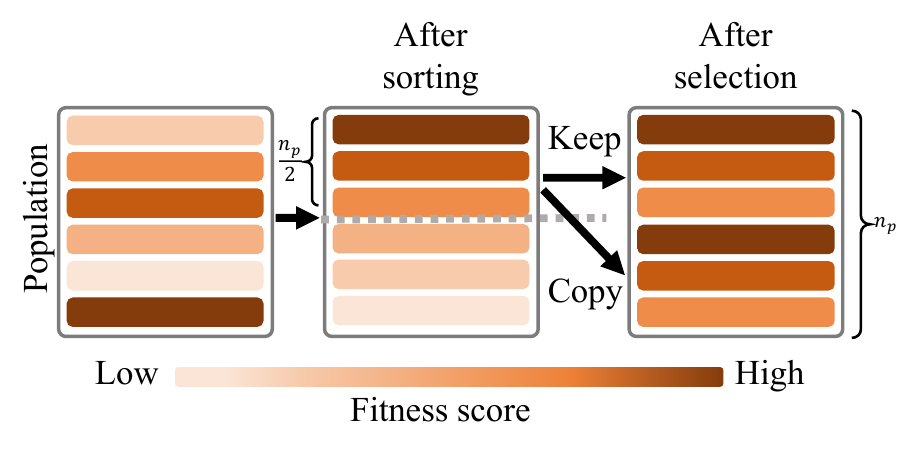}}
\caption{Illustration of the truncation-selection process. The scripts are sorted by their fitness scores, and then 50\% of the fittest scripts are selected and replicated twice.}
\label{fig:ga_selection}
\end{figure}

The selection step realizes the ``survival of the fittest.'' In other words, scripts with higher fitness scores are retained and replicated, whereas scripts with lower fitness scores are eliminated. In this study, the truncation selection method was used. Scripts were sorted by their fitness scores, and 50\% of the fittest scripts were selected and replicated twice. Figure~\ref{fig:ga_selection} shows the selection process.

\subsubsection{Crossover}

\begin{figure}[htbp!]
\centerline{\includegraphics[scale=1.]{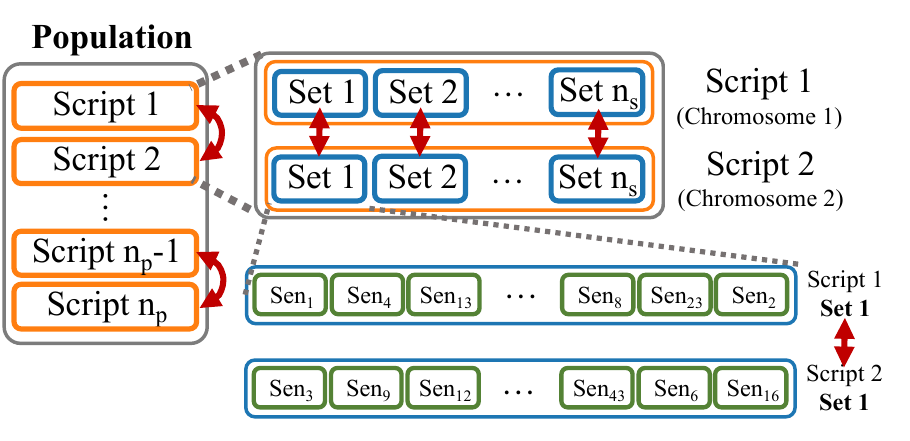}}
\caption{Illustration of the crossover pairs. The crossover step exchanges sentences between two sets with the same index.}
\label{fig:cross_pair}
\end{figure}

\begin{figure}[htbp!]
\centerline{\includegraphics[scale=.9]{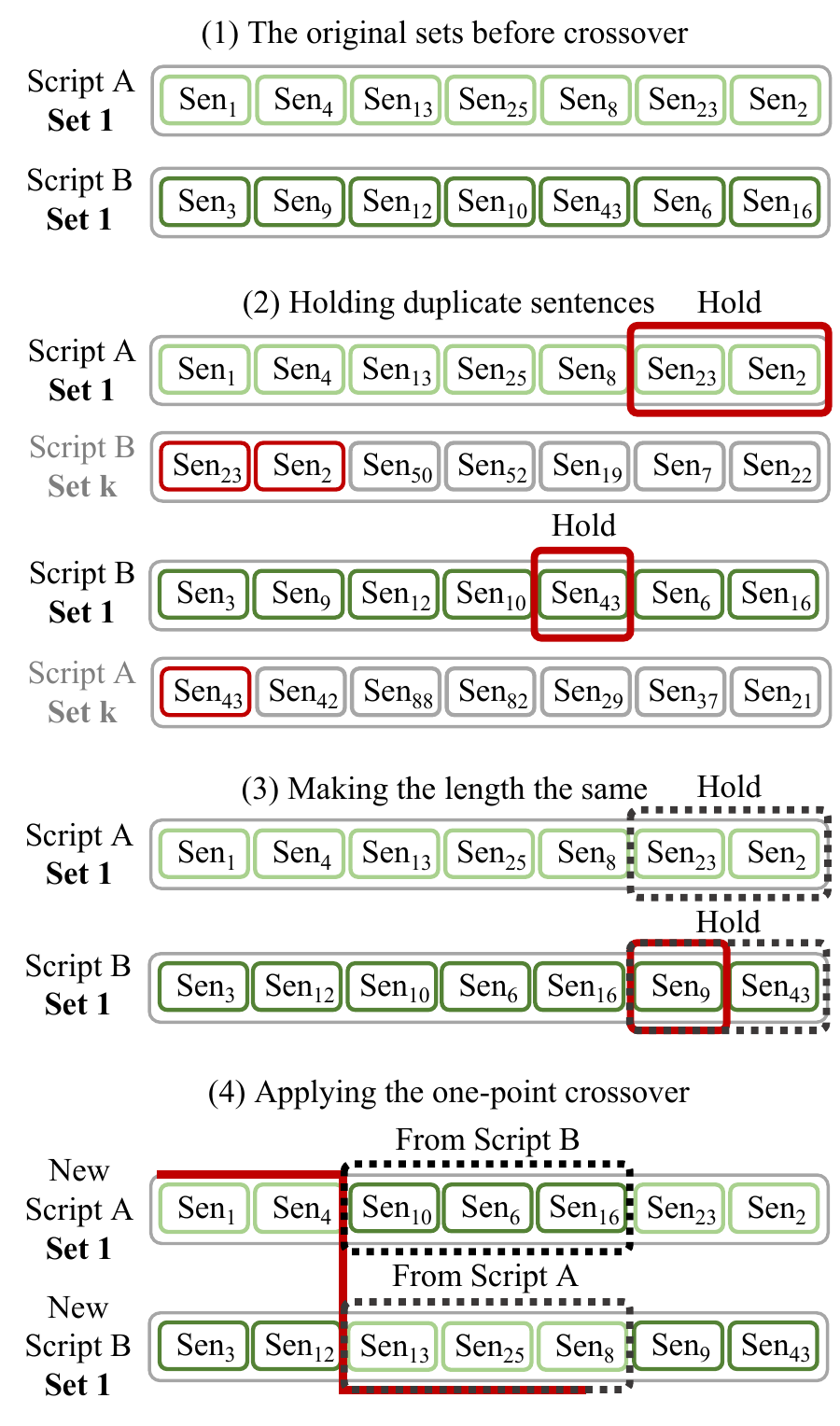}}
\caption{(1) The original sets before crossover. Sen$_i$ denotes the $i$-th sentence in the candidate sentence set. (2) Holding duplicate sentences. In this example, Sen$_{23}$ and Sen$_2$ are held because they exist in a set in Script B. Similarly, Sen$_{43}$ is held because it already exists a set in Script A. These sentences are not exchanged during the crossover process to avoid duplicate sentences in the script. If Sen$_{23}$ and Sen$_{2}$ are exchanged to Script B, there will be two Sen$_{23}$ and two Sen$_{2}$ in Script B. (3) Making the length the same. Because the number of duplicate sentences in Set 1 of Script A and Set 1 of Script B are not the same (i.e., two sentences in Set 1 of Script A and one sentence in Set 1 of Script B), we randomly hold one more sentence (Sen$_9$) in Set 1 of Script B. (4) Applying the one-point crossover.}
\label{fig:crossover_steps}
\end{figure}

The crossover step aims to combine the information of the two scripts and then generates new scripts. In this study, we used sets as crossover units, instead of complete scripts. This is because if we use scripts as crossover units, only one set in each script exchanges the information at every iteration when using the one-point crossover. However, if we use sets as crossover units, every set in the script participates in crossover at every iteration. Figure~\ref{fig:cross_pair} shows an example of a crossover pair and Figure~\ref{fig:crossover_steps} illustrates the crossover step. As shown in Figure~\ref{fig:crossover_steps}, to avoid duplicate sentences in one script, sentences present in the other script are held and not swapped in the crossover step. If the number of duplicate sentences in the paired sets is not the same, we randomly select sentences such that the number of held sentences is the same in both sets. Finally, we apply a one-point crossover to the two sets. Note that holding the same number of sentences in both sets ensures that the two new sets have the same number of sentences after crossover.

\subsection{Postprocessing}

After the script-composing phase, we obtained a syllable-balanced script. However, we may still want to replace some sentences in the script because the data-processing phase does not ensure that all candidate sentences are suitable. For example, the sensitive word filter cannot remove newly invented sensitive words that are not included in a sensitive word list. In addition, POS tagging systems may give incorrect POS tags because even the best POS tagging system cannot guarantee 100\% accuracy. Therefore, sentences that meet POS removal criteria may not be removed as expected. Moreover, sentences with low perplexity and high intelligibility scores are not necessarily logical from the human perspective.

Therefore, in the postprocessing phase, we still need to manually label inappropriate sentences to be replaced with more appropriate sentences. The script generated in the script-composing phase is denoted as a temporary script. We propose two methods to replace unwanted sentences in a temporary script: (1) GA-based method and (2) greedy-based method.

The GA-based method is similar to the GA in the script-composing phase. The only difference is the generation of scripts in the initial population. In postprocessing, all scripts in the initial population are initialized based on the temporary script, with unwanted sentences replaced with sentences randomly sampled from the candidate sentences. The rest of the GA steps were the same as those in the script-composing phase. For the greedy-based method, unwanted sentences are replaced one by one with sentences from the candidate sentences that can achieve the highest fitness score. According to our empirical results, the greedy-based method is more suitable when there are only a few unwanted sentences in the temporary script, whereas the GA-based method is more suitable when there are many unwanted sentences in the script.

\section{Experiments}
In this section, we first present a statistical analysis of Mandarin speech units based on Chinese news articles collected from five major news media outlets in Taiwan in 2021. We then show that the proposed BASPRO system can effectively select sentences based on a specially designed fitness function to form a syllable-balanced script for collecting speech data. Finally, we demonstrate that speech processing models trained on a TTS-synthesized syllable-balanced speech corpus based on the syllable-balanced script can achieve better performance than their counterparts trained on a randomly composed speech corpus. Note that the ``syllable distribution and coverage'' in the experiments represent ``\emph{tonal} syllable distribution and coverage''.

\subsection{Analysis of news articles in Taiwan in 2021}

We crawled news articles from five major news media sources in Taiwan in 2021, with a total Chinese character count of around 182,583,000. We used the Pypinyin tool \cite{ref_pypinyin} to identify the syllables of each character. See the Appendix for the list of INITIAL and FINAL in the Pypinyin tool, and the INITIAL, FINAL, and tone distribution in these news articles. There are 404 distinct base syllables and 1259 distinct tonal syllables, which are close to the number of distinct base syllables and tonal syllables reported in other studies \cite{wang1993auto, wang1998statistical}. Note that there is no consensus on the exact number of base and tonal syllables in Mandarin Chinese. For example, the number of base and tonal syllables in \cite{wang1998statistical} are 416 and 1345, respectively, while in \cite{wang1993auto} they are 407 and 1333, respectively.

\subsection{Data processing experiment}

\subsubsection{Experimental settings of data processing}
The general filter kept only ten-character sentences. The POS tagging filter removes sentences that satisfy the POS-based removal criteria using CkipTagger\cite{li2020attention} or DDParser\cite{zhang2020practical}. The removal criteria when using the CkipTagger and DDParser are listed in Table~\ref{table:postagging_criteria}. The perplexity filter removes sentences with perplexities higher than 4.0. In intelligibility filter, only sentences with an intelligibility score of 1.0 were kept. After the data-processing phase, the total number of candidate sentences was around 167,000. Table~\ref{table:process_toolkits} lists the toolkits used in each data-processing phase.  

\begin{table}[htbp!]
\centering
\caption{The POS-based Removal Criteria. Descriptions of POS tags can be found in \cite{li2020attention} and \cite{zhang2020practical}}
\label{table:postagging_criteria}
\resizebox{\columnwidth}{!}{%
\begin{tabular}{|c|c|c|c|}
\hline
Toolkit& Include& Start& End                 \\ \hline
CkipTagger\cite{li2020attention} & 'Nb','Nc','FW' & 'DE','SHI','T' & \begin{tabular}[c]{@{}c@{}}'Caa','Cab','Cba',\\ 'Cbb','P','T'\end{tabular} \\ \hline
DDParser\cite{zhang2020practical} & \begin{tabular}[c]{@{}c@{}}'LOC','ORG','TIME',\\ 'PER','w','nz'\end{tabular} & 'p','u','c'    & 'xc','u' \\ \hline
\end{tabular}%
}
\end{table}

\begin{table}[htbp!]
\centering
\caption{Data processing toolkits used in this study}
\label{table:process_toolkits}
\resizebox{\columnwidth}{!}{%
\begin{tabular}{|c|c|c|c|}
\hline
\multirow{2}{*}{\begin{tabular}[c]{@{}c@{}}POS \\ filter\end{tabular}} & \multirow{2}{*}{\begin{tabular}[c]{@{}c@{}}Perplexity \\ filter\end{tabular}} & \multirow{2}{*}{\begin{tabular}[c]{@{}c@{}}Intelligibility \\ filter\end{tabular}} & \multirow{2}{*}{\begin{tabular}[c]{@{}c@{}}Syllable \\ calculation\end{tabular}} \\ & & & \\ \hline
\begin{tabular}[c]{@{}c@{}}CkipTagger\cite{li2020attention}\\ DDParser\cite{zhang2020practical}\end{tabular}          & \begin{tabular}[c]{@{}c@{}}Hugging Face\cite{wolf2019huggingface}\\ (bert-base-chinese)\end{tabular}    & \begin{tabular}[c]{@{}c@{}}Google-TTS \cite{ref_google_tts}\\ Google-ASR\cite{ref_google_asr}\end{tabular}& Pypinyin\cite{ref_pypinyin}\\ \hline
\end{tabular}%
}
\end{table}

\subsubsection{Experimental results of data processing}

\begin{figure}[htbp!]
\centerline{\includegraphics[scale=0.9]{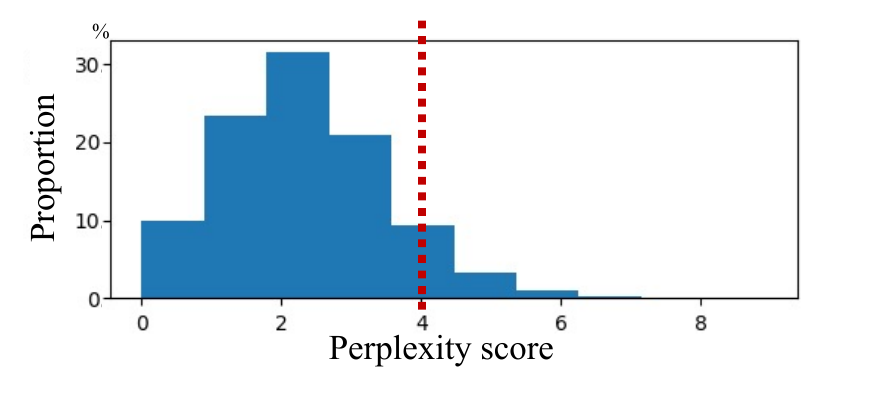}}
\caption{Perplexity distribution for ten-character sentences in Mandarin Chinese news texts. The red dotted line represents the threshold used for the perplexity filter.}
\label{fig:perplexity_distribution}
\end{figure}

Table~\ref{table:perplexity_filter} lists several examples of sentences and their corresponding perplexities. The experimental results showed that perplexity can reflect human perception to a certain extent. Specifically, sentences 1-1, 2-1, and 3-1 are literally similar to sentences 1-2, 2-2, and 3-2, respectively. Only a few characters in each sentence pair were different, and the pronunciations of the different characters were similar. However, sentences 1-1, 2-1, 3-1 are considered natural, while sentences 1-2, 2-2, 3-2 contain typos or are illogical. According to the results in Table~\ref{table:perplexity_filter}, sentences 1-2, 2-2, 3-2 have higher perplexity, while sentences 1-1, 2-1, 3-1 have lower perplexity. Figure~\ref{fig:perplexity_distribution} shows the perplexity distribution for ten-character sentences in Mandarin Chinese news texts. The distribution of perplexity was right-skewed, with a mean of 2.336. According to Figure~\ref{fig:perplexity_distribution}, we chose 4.0 as the perplexity threshold, which is approximately 1.5 standard deviations from the mean of perplexity for all ten-character sentences. However, sometimes the perplexity does not correctly reflect whether a sentence is understandable. For example, sentence 4 in Table~\ref{table:perplexity_filter} is difficult to understand but has the lowest perplexity among the examples.

\begin{table}[htbp!]
\centering
\caption{Examples of sentence perplexity assessment}
\label{table:perplexity_filter}
\includegraphics[scale=0.9]{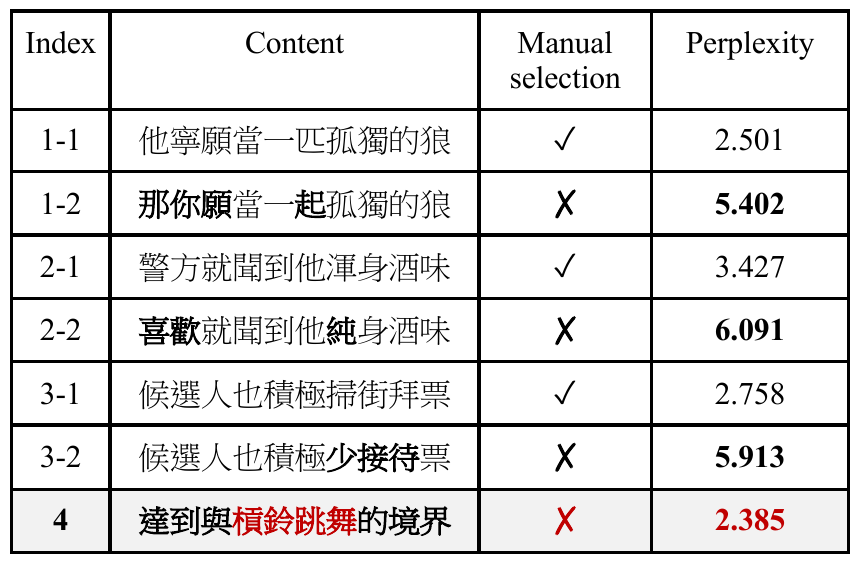}
\end{table}

Table~\ref{table:intell_filter} lists examples of sentences and their corresponding intelligibility scores. ``Ori'' is the original input sentence, and ``Pred'' is the corresponding ASR prediction. The first and second examples show that the intelligibility filter can identify sentences with words that are not easy to understand. To avoid the need to replace many sentences in the postprocessing phase, the intelligibility filter removes all sentences with an intelligibility score lower than 1.0. In other words, the intelligibility filter only retained sentences with perfect ASR test results. However, like perplexity, sometimes, the intelligibility score does not perfectly reflect human perception. For example, the third sentence is not intuitive but has the intelligibility score of 1.0. As shown in Tables~\ref{table:perplexity_filter} and \ref{table:intell_filter}, perplexity and intelligibility filters cannot remove all illogical sentences. Therefore, manual labeling is required during the postprocessing phase.

\begin{table}[htbp!]
\centering
\caption{Examples of sentence intelligibility assessment}
\label{table:intell_filter}
\includegraphics[scale=.8]{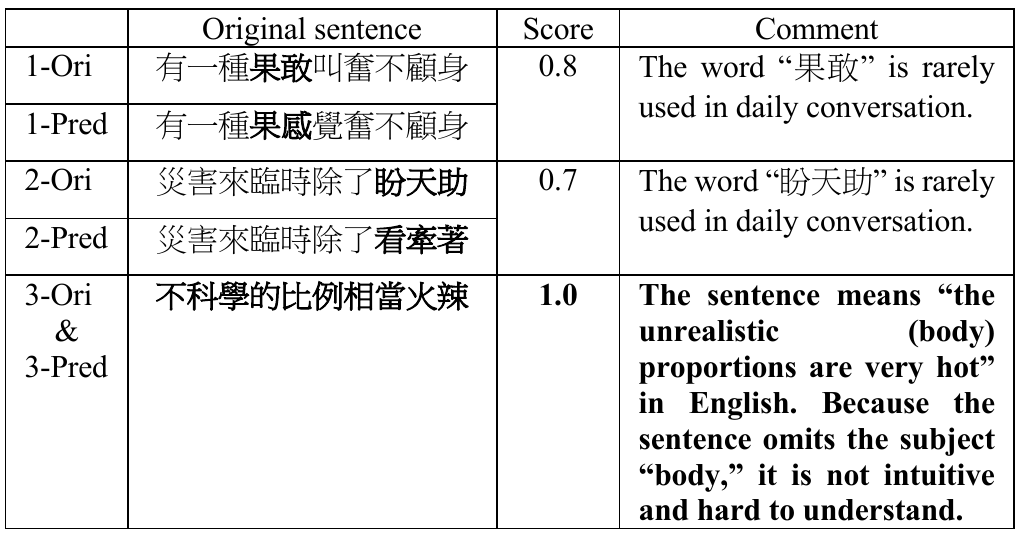}
\end{table}

\subsection{GA-based script-composing experiment}

\begin{figure}[htbp!]
\centerline{\includegraphics[scale=.9]{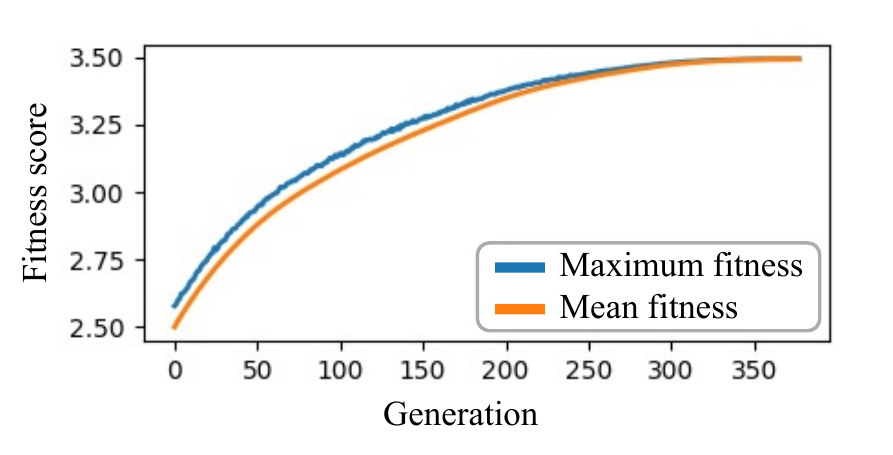}}
\caption{Training curve of the GA. Overall, the fitness score increases with the number of generations and then eventually converges.}
\label{fig:ga_training_curve}
\end{figure}

In this section, we demonstrate that the BASPRO system can effectively select sentences to form a recording script according to the designed fitness function. We set the number of sets in the script and the number of sentences in each set to 20. Thus, the length of the chromosomes was 400. The weight of script\_syllable\_coverage ($w_2$ in Eq.~\ref{eq:fitness}) was set to two, whereas the weights of the script\_syllable\_distribution ($w_1$ in Eq.~\ref{eq:fitness}) and set\_syllable\_distribution ($w_3$ in Eq.~\ref{eq:fitness}) was set to 1. The population size was set to 25,000 and the GA was stopped until the maximum fitness score converged. Figure~\ref{fig:ga_training_curve} shows the training curve of GA. The maximum fitness score drops for some generations because scripts are split and remixed in the crossover step, which may lower the fitness score. However, overall, the fitness score increases with the number of generations and eventually converges.

\begin{figure}[htbp!]
\centerline{\includegraphics[scale=1.]{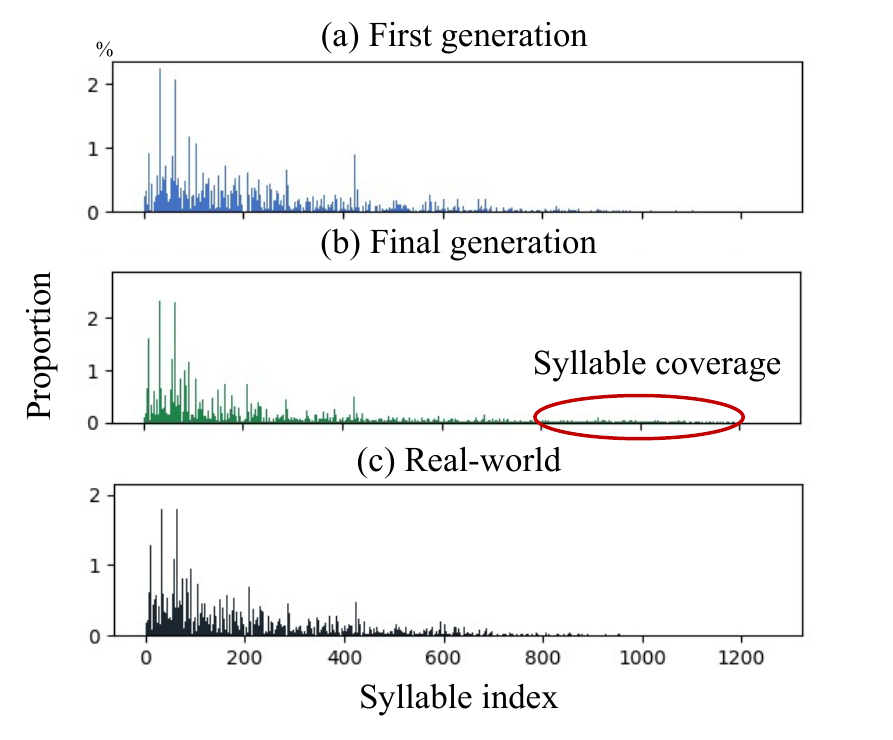}}
\caption{Distribution of syllables in the best scripts of the first and final generations and in real-world texts. The results show that the best script in the final generation has a syllable distribution that is closer to real-world syllable distribution than the best script in the first generation. The red region reveals the effect of script\_syllable\_coverage; that is, more rare syllables are covered in the best script in the final generation.}
\label{fig:ga_training_distribution}
\end{figure}

Figure~\ref{fig:ga_training_distribution} shows the distribution of syllables in the best scripts of the first and final generations, and in real-world texts. The results showed that the syllable distribution of the best script in the final generation was much closer to the real-world syllable distribution than the syllable distribution of the best script in the first generation. The red region in Figure~\ref{fig:ga_training_distribution} indicates the effect of script\_syllable\_coverage score on the fitness function. In the real world, the ratio of the frequency of syllables with indices 800 to 1200 to the frequency of all syllables is close to 0; therefore, when considering only script\_syllable\_distribution and set\_syllable\_distribution in the fitness function, most syllables in this rare region will not be present in the best script in the final generation. However, because the fitness function includes script\_syllable\_coverage, more rare syllables are covered in the best script in the final generation, making the distribution of syllables indexed from 800 to 1200 in (b) and (c) significantly different.

Table~\ref{table:ga_training_scores} compares the values of script\_syllable\_distribution, set\_syllable\_distribution, and script\_syllable\_coverage for the best scripts in the first and final generations. Note that because there were 20 sets in a script, for the set\_syllable\_distribution, the mean and standard deviation of the 20 sets were calculated. Clearly, all values increase with generation. As shown in the ablation study in Table~\ref{table:ga_compare_fitness}, there is a tradeoff between script\_syllable\_distribution, set\_syllable\_distribution, and syllable\_coverage. For example, if the fitness function only considers the script\_syllable\_distribution, the best final script can achieve a script\_syllable\_distribution value of 0.997. However, in this case, the script\_syllable\_coverage and set\_syllable\_distribution can only reach 579 and 0.702, respectively. 

\begin{table}[htbp!]
\centering
\caption{Statistics of the best scripts in the first and final generations}
\label{table:ga_training_scores}
\resizebox{\columnwidth}{!}{%
\begin{tabular}{|c|c|cc|}
\hline& & \multicolumn{2}{c|}{Syllable distribution}                              \\ \cline{3-4} 
\multirow{-2}{*}{Generation} & \multirow{-2}{*}{\begin{tabular}[c]{@{}c@{}}Syllable \\ coverage\end{tabular}} & \multicolumn{1}{c|}{Script} & Set \\ \hline
First & 668  & \multicolumn{1}{c|}{0.894}                         & 0.622 (std: 0.033) \\ \hline
\rowcolor[HTML]{EFEFEF} 
Final & 1120 & \multicolumn{1}{c|}{\cellcolor[HTML]{EFEFEF}0.964} & 0.751 (std: 0.019)  \\ \hline
\end{tabular}%
}
\end{table}

\begin{table}[htbp!]
\centering
\caption{Ablation study of the fitness function}
\label{table:ga_compare_fitness}
\resizebox{\columnwidth}{!}{%
\begin{tabular}{|c|c|cc|}
\hline
&                                    & \multicolumn{2}{c|}{Syllable distribution}        \\ \cline{3-4} 
\multirow{-2}{*}{\begin{tabular}[c]{@{}c@{}}Fitness \\ function\end{tabular}} &
  \multirow{-2}{*}{\begin{tabular}[c]{@{}c@{}}Syllable \\ coverage\end{tabular}} &
  \multicolumn{1}{c|}{Script} &
  Set \\ \hline \rowcolor[HTML]{EFEFEF}
All  & 1120& \multicolumn{1}{c|}{0.964}             & 0.751(std:0.019) \\ \hline
Syllable coverage   & {\color[HTML]{CB0000} \textbf{1122}} & \multicolumn{1}{c|}{0.827}                                 & 0.494(std:0.035)  \\ \hline
Script distribution & 579                                  & \multicolumn{1}{c|}{{\color[HTML]{CB0000} \textbf{0.997}}} & 0.702(std:0.042)                                 \\ \hline
Set distribution    & 343                                  & \multicolumn{1}{c|}{0.943}                                 & {\color[HTML]{CB0000} \textbf{0.889(std:0.003)}} \\ \hline
\end{tabular}%
}
\end{table}

\begin{figure}[htbp!]
\centerline{\includegraphics[scale=1.]{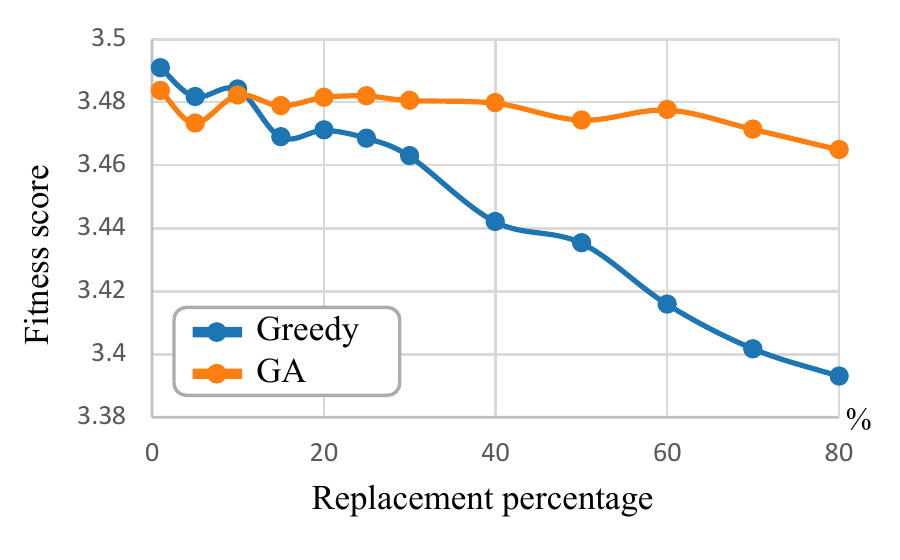}}
\caption{Comparison between the GA- and greedy-based replacement methods in the postprocessing phase. The greedy-based method outperforms the GA-based method when the replacement percentage is lower than 10\%; however, as the replacement percentage increases, the GA-based method outperforms the greedy-based method.}
\label{fig:greedy_ga_compare}
\end{figure}

Next, we compare the greedy and GA-based replacement methods in the postprocessing phase. Figure~\ref{fig:greedy_ga_compare} shows the fitness scores of the resulting scripts for different replacement percentages. Specifically, 80\% means that 320 (i.e., 400 $\times$ 0.8) sentences in the script have been replaced with new sentences. The results show that if a large portion of sentences needs to be replaced, the GA-based method performs better than the greedy-based method. Conversely, if only a few sentences must be replaced, the greedy method outperforms the GA-based method. 

\begin{figure}[htbp!]
\centerline{\includegraphics[scale=1.]{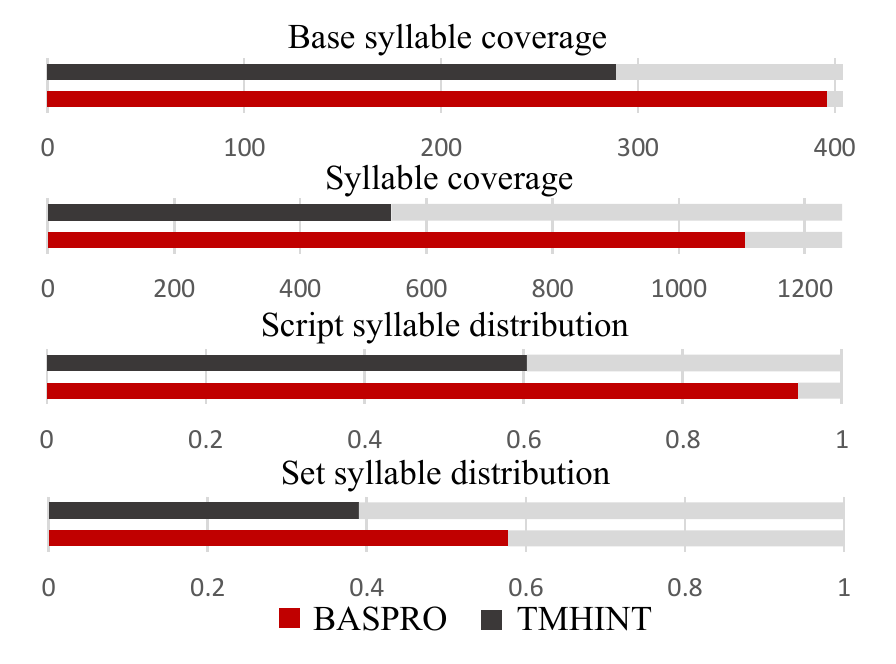}}
\caption{Comparison between the script produced by the BASPRO system and TMHINT script. The maximum base syllable and (tonal) syllable coverage were 404 and 1259, respectively. The maximum corpus syllable distribution and syllable distribution scores are 1.}
\label{fig:tmhit_compare}
\end{figure}

Finally, Figure~\ref{fig:tmhit_compare} compares the statistics of a script produced by the BASPRO system and the TMHINT Mandarin Chinese recording script \cite{huang2005development} used in many previous studies. For a fair comparison, the number of sets and sentences in each set was set to 32 and 10, respectively, following the TMHINT script. The top two panels of Figure~\ref{fig:tmhit_compare} show that the BASPRO-produced script covers more syllables, while the bottom two panels of Figure~\ref{fig:tmhit_compare} show that the syllable distribution of the BASPRO-produced script is closer to the real-world syllable distribution.

\subsection{Experiment on speech-processing tasks}
In this section, we investigate whether speech-processing models trained on the syllable-balanced \emph{TMNews} corpus can outperform their counterparts trained on a randomly composed corpus. We experiment on two common speech processing tasks, including speech enhancement (SE) and ASR.

\subsubsection{Experimental settings for both tasks}

To verify the usefulness of the proposed BASPRO system, we compared the performances of speech-processing models trained on syllable-balanced and randomly selected corpora. In the following experiments, CorpusBAL referred to a syllable-balanced corpus, whereas CorpusRAN represented a randomly composed corpus. CorpusBAL was formed based on a syllable-balanced script, \emph{TMNews}. CorpusRAN was formed using randomly selected sentences. Both CorpusBAL and CorpusRAN have large and small versions, denoted by Corpus(BAL,RAN)\_{\footnotesize Large} and Corpus(BAL,RAN)\_{\footnotesize Small}, respectively. The large and small corpora contained 20 and 5 sets, respectively, with 20 sentences in each set. That is, 400 sentences form a large corpus and 100 sentences form a small corpus.

For each sentence in the script, we used two TTS systems, GoogleTTS \cite{ref_google_tts} and TTSkit \cite{ref_ttskit}, to generate corresponding utterances. The utterances generated by GoogleTTS were female voices, while the utterances generated by TTSkit were male voices. As a result, the Corpus(BAL,RAN)\_{\footnotesize Large} corpus contains 800 utterances, and the Corpus(BAL,RAN)\_{\footnotesize Small} corpus contains 200 utterances.

Table~\ref{table:corpus_statistic} lists the statistics of each speech corpus. The syllable distribution of CorpusBAL was closer to the real-world syllable distribution than that of CorpusRAN. In addition, CorpusBAL\_{\footnotesize Small} had better syllable coverage than CorpusRAN\_{\footnotesize Large}, although the number of sentences in CorpusBAL\_{\footnotesize Small} was only a quarter of that in CorpusRAN\_{\footnotesize Large}.    

\begin{table}[htbp!]
\centering
\caption{Statistics of the speech corpora}
\label{table:corpus_statistic}
\resizebox{\columnwidth}{!}{%
\begin{tabular}{|c|c|c|c|c|}
\hline
Corpus &
  \begin{tabular}[c]{@{}c@{}}Base\\ syllable\\ coverage\end{tabular} &
  \begin{tabular}[c]{@{}c@{}}Syllable\\ coverage\end{tabular} &
  \begin{tabular}[c]{@{}c@{}}Script\\ syllable\\ distribution\end{tabular} &
  \begin{tabular}[c]{@{}c@{}}Set\\ syllable\\ distribution\end{tabular} \\ \hline
\rowcolor[HTML]{EFEFEF}\begin{tabular}[c]{@{}c@{}}CorpusBAL\_{\scriptsize Large} \\ (TMNews\_{\scriptsize L})\end{tabular}& 392 & 1061 & 0.970 & \begin{tabular}[c]{@{}c@{}}0.743\\ (std: 0.020)\end{tabular}  \\ \hline
\rowcolor[HTML]{EFEFEF} 
\begin{tabular}[c]{@{}c@{}}CorpusBAL\_{\scriptsize Small} \\ (TMNews\_{\scriptsize S}) \end{tabular}& 333 & 629  & 0.934 & \begin{tabular}[c]{@{}c@{}}0.701\\ (std: 0.10)\end{tabular}   \\ \hline
CorpusRAN\_{\scriptsize Large} & 319 & 609  & 0.869 & \begin{tabular}[c]{@{}c@{}}0.603\\ (std: 0.365)\end{tabular} \\ \hline
CorpusRAN\_{\scriptsize Small} & 241 & 387  & 0.818 & \begin{tabular}[c]{@{}c@{}}0.637\\ (std: 0.015)\end{tabular}  \\ \hline
\end{tabular}%
}
\end{table}

\paragraph{Experimental settings for the SE task}
We trained the SE model on small corpora, and tested it on large corpora. In practical applications, the test data are also larger than the training data. Therefore, we believe that the experimental results under this setting can better reflect performance in a real environment.

For the training data, each clean utterance was contaminated with 25 noises randomly selected from 100 noises \cite{hu2004100} at -1, 1, 3, and 5 SNR levels. The training data contained 20,000 utterances (100 (sentences) $\times$ 2 (voice types) $\times$ 25 (noise types) $\times$ 4 (SNR levels)). The training data were divided into training and validation datasets. The validation set contained 20\% of the training data and was used to select the best model for training. Therefore, in our experiments, using this training–validation setup, we trained five models with a training corpus and reported the mean and standard deviation of the results evaluated on the testing corpus. For the test set, each clean utterance was contaminated with three noise types (white, street, and babble) at 2 and 4 SNR levels. The test set contains 4,800 utterances (400 (sentences) $\times$ 2 (voice types) $\times$ 3 (noise types) $\times$ 2 (SNR levels)).

The corpora were evaluated using MetricGAN+\cite{fu2019metricgan, fu2021metricgan+}, a state-of-the-art SE model. Because the input of MetricGAN+ is a spectrogram, the input signal was transformed into a spectrogram using a short-time Fourier transform (STFT) with a window length of 512 and hop length of 256. In addition, the batch size was 32, the loss function used was L1 loss, and the optimizer was Adam with a learning rate of 0.001. The perceptual evaluation of speech quality (PESQ) \cite{rix2001perceptual} and short-time objective intelligibility (STOI) \cite{taal2011algorithm} are used as objective evaluation metrics.  

\paragraph{Experimental settings for the ASR task}
In the ASR experiments, we downloaded the pretrained transformer-based ASR model from SpeechBrain \cite{speechbrain}, and then fine-tuned the ASR model using the speech corpora collected in this study. The pre-trained ASR model was trained on the AISHELL dataset, which is also a Mandarin speech corpus. We fine-tuned a pre-trained model because our training speech was not sufficient to train the ASR model from scratch. In addition, this setup simulates the personalization of an ASR system, that is, fine-tuning an ASR system with a few recordings of a new user. Similar to the 80\% training-20\% validation setting in the SE task, given a training corpus, we obtained five models and reported the means and standard deviations of the evaluation results. For each training and validation split, we fine-tuned the model for 50 epochs and selected the best model using a validation set.

We used the pinyin error rate (PER), character error rate (CER), and sentence error rate (SER) to evaluate ASR performance. PER calculates the difference between the predicted and ground-truth syllable sequences. Note that Pypinyin \cite{ref_pypinyin} was used to convert characters to tonal syllables before calculating PER. PER and CER were calculated using Levenshtein distance. In SER, a predicted sentence is considered to be incorrect if any character is wrong.

\subsubsection{Experimental results for SE}

Table~\ref{table:SE_results} compares the performances of the SE models trained on CorpusBAL\_{\footnotesize Small} and CorpusRAN\_{\footnotesize Small}. 
The results show that the SE model trained on CorpusBAL\_{\footnotesize Small} outperformed the SE model trained on CorpusRAN\_{\footnotesize Small} in terms of both PESQ and STOI under all testing conditions. In addition, both models performed worse when tested on CorpusBAL\_{\footnotesize Large} than on CorpusRAN\_{\footnotesize Large}. This may be because CorpusBAL\_{\footnotesize Large} covers more syllables than CorpusRAN\_{\footnotesize Large}, thus making it a more challenging test corpus. Table~\ref{table:SE_results_ttest} presents the corresponding t-test results. The p-values of the STOI results on both CorpusBAL\_{\footnotesize Large} and CorpusRAN\_{\footnotesize Large} testing data are about 0.1, while the p-values of the PESQ results are about 0.5. That is, the improvement in the SE performance on STOI is more statistically significant than that on PESQ. This result may be because syllable coverage and distribution have a greater impact on intelligibility (STOI) than on quality (PESQ).

\begin{table}[htbp!]
\centering
\caption{Performance of the SE models trained on CorpusBAL and CorpusRAN}
\label{table:SE_results}
\resizebox{\columnwidth}{!}{%
\begin{tabular}{|c|c|c|c|c|} 
\hline
\diagbox{Testing}{Training} & \multicolumn{2}{c|}{{\cellcolor[HTML]{EFEFEF}}CorpusBAL\_{\scriptsize Small}}  & \multicolumn{2}{c|}{CorpusRAN\_{\scriptsize Small}}  \\ 
\hline & {\cellcolor[HTML]{EFEFEF}}STOI  & {\cellcolor[HTML]{EFEFEF}}PESQ & STOI  & PESQ \\ 
\hline
CorpusBAL\_{\scriptsize Large} & {\cellcolor[HTML]{EFEFEF}}\begin{tabular}[c]{@{}>{\cellcolor[HTML]{EFEFEF}}c@{}}\textbf{0.832}\\(std: 0.0149)\end{tabular} & {\cellcolor[HTML]{EFEFEF}}\begin{tabular}[c]{@{}>{\cellcolor[HTML]{EFEFEF}}c@{}}\textbf{1.792}\\(std: 0.1154)\end{tabular} & \begin{tabular}[c]{@{}c@{}}0.793\\(std: 0.0426)\end{tabular} & \begin{tabular}[c]{@{}c@{}}1.744\\(std: 0.1068)\end{tabular}  \\ 
\hline
CorpusRAN\_{\scriptsize Large}                   & {\cellcolor[HTML]{EFEFEF}}\begin{tabular}[c]{@{}>{\cellcolor[HTML]{EFEFEF}}c@{}}\textbf{0.832}\\(std: 0.0133)\end{tabular} & {\cellcolor[HTML]{EFEFEF}}\begin{tabular}[c]{@{}>{\cellcolor[HTML]{EFEFEF}}c@{}}\textbf{1.804}\\(std: 0.1182)\end{tabular} & \begin{tabular}[c]{@{}c@{}}0.796\\(std: 0.0426)\end{tabular} & \begin{tabular}[c]{@{}c@{}}1.755\\(std: 0.1101)\end{tabular}   \\
\hline
\end{tabular}%
}
\end{table}

\begin{table}[htbp!]
\centering
\caption{T-test of the CorpusBAL\_{\scriptsize Small} and CorpusRAN\_{\scriptsize Small} SE results}
\label{table:SE_results_ttest}
\begin{tabular}{|c|cc|}
\hline
& \multicolumn{2}{c|}{p-value}           \\ \hline
Testing data     & \multicolumn{1}{c|}{STOI}    & PESQ    \\ \hline
CorpusBAL\_{\scriptsize Large} & \multicolumn{1}{c|}{0.10028} & 0.51936 \\ \hline
CorpusRAN\_{\scriptsize Large}& \multicolumn{1}{c|}{0.10524} & 0.46861 \\ \hline
\end{tabular}
\end{table}

The fitness function contains the set\_syllable\_distribution score, because we want each set to be representative. We argue that the model selected by a small syllable-balanced validation set is more robust than the model selected by a small randomly selected validation set. Table~\ref{table:SE_validation} compares the performances of the SE models selected with different validation sets. The SE model was trained on CorpusBAL\_{\footnotesize Small} and tested on CorpusBAL\_{\footnotesize Large} and CorpusRAN\_{\footnotesize Large}. In Table~\ref{table:SE_validation}, valid:bal indicates that the validation set is a syllable-balanced set in CorpusBAL\_{\footnotesize Small}, whereas valid:ran indicates that the validation set is randomly selected sentences from CorpusBAL\_{\footnotesize Small}. The results show that the average performance of the SE models selected with a syllable-balanced validation set is better than that of the SE models selected with a randomly selected validation set.

\begin{table}[htbp!]
\centering
\caption{SE performance using different validation sets}
\label{table:SE_validation}
\resizebox{\columnwidth}{!}{%
\begin{tabular}{|c|c|c|c|c|} 
\hline
\diagbox{Testing}{Training} & \multicolumn{2}{c|}{{\cellcolor[HTML]{EFEFEF}}\begin{tabular}[c]{@{}>{\cellcolor[HTML]{EFEFEF}}c@{}}CorpusBAL\_{\scriptsize Small}\\(valid:bal)\end{tabular}} & \multicolumn{2}{c|}{\begin{tabular}[c]{@{}c@{}}CorpusBAL\_{\scriptsize Small}\\(valid:ran)\end{tabular}}                                        \\ 
\hline & {\cellcolor[HTML]{EFEFEF}}STOI   & {\cellcolor[HTML]{EFEFEF}}PESQ  & STOI  & PESQ \\ 
\hline 
CorpusBAL\_{\scriptsize Large} & {\cellcolor[HTML]{EFEFEF}}\begin{tabular}[c]{@{}>{\cellcolor[HTML]{EFEFEF}}c@{}}\textbf{0.832}\\(std: 0.0149)\end{tabular} & {\cellcolor[HTML]{EFEFEF}}\begin{tabular}[c]{@{}>{\cellcolor[HTML]{EFEFEF}}c@{}}\textbf{1.792}\\(std: 0.1154)\end{tabular} & \begin{tabular}[c]{@{}c@{}}0.814\\(std: 0.0266)\end{tabular} & \begin{tabular}[c]{@{}c@{}}1.790\\(std: 0.0655)\end{tabular}  \\ 
\hline
CorpusRAN\_{\scriptsize Large}                 & {\cellcolor[HTML]{EFEFEF}}\begin{tabular}[c]{@{}>{\cellcolor[HTML]{EFEFEF}}c@{}}\textbf{0.832}\\(std: 0.0133)\end{tabular} & {\cellcolor[HTML]{EFEFEF}}\begin{tabular}[c]{@{}>{\cellcolor[HTML]{EFEFEF}}c@{}}\textbf{1.804}\\(std: 0.1182)\end{tabular} & \begin{tabular}[c]{@{}c@{}}0.816\\(std: 0.0265)\end{tabular} & \begin{tabular}[c]{@{}c@{}}1.802\\(std: 0.0694)\end{tabular}  \\
\hline
\end{tabular}%
}
\end{table}

\subsubsection{Experimental results for ASR}

Table~\ref{table:ASR_results} shows the performance of the ASR models fine-tuned using CorpusBAL and CorpusRAN. First, the results reveal that fine-tuning an ASR model always improves ASR performance. In addition, the ASR models fine-tuned on CorpusBAL generally performed better than their corresponding models fine-tuned on CorpusRAN. This is because the CorpusBAL\_{\footnotesize Large} and CorpusBAL\_{\footnotesize Small} corpora cover relatively complete and rich pronunciations; thus, the ASR model can be fine-tuned comprehensively. However, we also see that when tested on CorpusRAN\_{\footnotesize Small}, the ASR model fine-tuned on CorpusBAL\_{\footnotesize Large} performs slightly worse than the ASR model fine-tuned on CorpusRAN\_{\footnotesize Large}. One possible explanation is that both CorpusBAL\_{\footnotesize Large} and CorpusRAN\_{\footnotesize Large} cover more syllables than CorpusRAN\_{\footnotesize Small}, as shown in Table~\ref{table:corpus_statistic}. Therefore, fine-tuning the model with either corpus did not make a significant difference when testing on a small test set. However, such a biased small test set could mislead the model. When using a small corpus as a test set, more consideration should be given to the pronunciation balance and coverage. Finally, the ASR performance tested on CorpusRAN is better than the ASR performance tested on CorpusBAL, which is consistent with the SE experiments. This is because CorpusBAL covers more rare syllables and is, therefore, more challenging than CorpusRAN. 

Table~\ref{table:ASR_results_ttest} presents the corresponding t-test results. This evaluation shows that the performance of the two ASR models using corpora of different scripts across all evaluation metrics is significantly different on the CorpusBAL\_{\footnotesize Large} and CorpusBAL\_{\footnotesize Small} testing data (p-value$\ll$0.05). On the CorpusRAN\_{\footnotesize Large} testing data, the p-value for CER is 0.18967, which means that the performance difference is not significant. Note that the CER is the only case in which CorpusRAN\_{\footnotesize Small} performs better than CorpusBAL\_{\footnotesize Small} on CorpusRAN\_{\footnotesize Large} in Table~\ref{table:ASR_results}. On the CorpusRAN\_{\footnotesize Small} testing data, the performance differences in PER, CER, and SER are not significant (p-value\textgreater 0.05). The experimental results show that syllable coverage and distribution should be considered for both training data and testing data, especially when the amount of data is small.

\begin{table}[htbp!]
\centering
\caption{Performance of ASR models trained on CorpusBAL and CorpusRAN}
\label{table:ASR_results}
\resizebox{\columnwidth}{!}{%
\begin{tabular}{|c|c|c|c|c|}
\hline
Testing data &
  Training data &
  PER &
  CER &
  SER \\ \hline
 &
  w/o fine-tuned &
  14.94 &
  19.73 &
  74.88 \\ \cline{2-5} 
 &
  \cellcolor[HTML]{EFEFEF}CorpusBAL\_{\scriptsize Small} &
  \cellcolor[HTML]{EFEFEF}\begin{tabular}[c]{@{}c@{}}\textbf{9.658}   \\ (std: 0.259)\end{tabular} &
  \cellcolor[HTML]{EFEFEF}\begin{tabular}[c]{@{}c@{}}\textbf{15.544} \\ (std: 0.212)\end{tabular} &
  \cellcolor[HTML]{EFEFEF}\begin{tabular}[c]{@{}c@{}}\textbf{67.648} \\ (std: 0.957)\end{tabular} \\ \cline{2-5} 
\multirow{-3}{*}{CorpusBAL\_{\scriptsize Large}} &
  CorpusRAN\_{\scriptsize Small} &
  \begin{tabular}[c]{@{}c@{}}10.738 \\ (std: 0.277)\end{tabular} &
  \begin{tabular}[c]{@{}c@{}}16.696 \\ (std: 0.264)\end{tabular} &
  \begin{tabular}[c]{@{}c@{}}70.324\\ (std: 1.311)\end{tabular} \\ \hline
 &
  w/o fine-tuned &
  8.78 &
  11.69 &
  55.62 \\ \cline{2-5} 
 &
  \cellcolor[HTML]{EFEFEF}CorpusBAL\_{\scriptsize Small} &
  \cellcolor[HTML]{EFEFEF}\begin{tabular}[c]{@{}c@{}}\textbf{4.885} \\ (std: 0.062)\end{tabular} &
  \cellcolor[HTML]{EFEFEF}\begin{tabular}[c]{@{}c@{}}9.244 \\ (std: 0.141)\end{tabular} &
  \cellcolor[HTML]{EFEFEF}\begin{tabular}[c]{@{}c@{}}\textbf{47.922} \\ (std: 0.518)\end{tabular} \\ \cline{2-5} 
\multirow{-3}{*}{CorpusRAN\_{\scriptsize Large}} &
  CorpusRAN\_{\scriptsize Small} &
  \begin{tabular}[c]{@{}c@{}}5.063 \\ (std: 0.034)\end{tabular} &
  \begin{tabular}[c]{@{}c@{}}\textbf{9.094} \\ (std: 0.186)\end{tabular} &
  \begin{tabular}[c]{@{}c@{}}49.126 \\ (std: 0.905)\end{tabular} \\ \hline \hline
 &
  w/o fine-tuned &
  14.30 &
  17.75 &
  70.00 \\ \cline{2-5} 
 &
  \cellcolor[HTML]{EFEFEF}CorpusBAL\_{\scriptsize Large} &
  \cellcolor[HTML]{EFEFEF}\begin{tabular}[c]{@{}c@{}}\textbf{6.61}\\  (std: 0.163)\end{tabular} &
  \cellcolor[HTML]{EFEFEF}\begin{tabular}[c]{@{}c@{}}\textbf{11.84}\\ (std: 0.397)\end{tabular} &
  \cellcolor[HTML]{EFEFEF}\begin{tabular}[c]{@{}c@{}}\textbf{56.70} \\ (std: 1.823)\end{tabular} \\ \cline{2-5} 
\multirow{-3}{*}{CorpusBAL\_{\scriptsize Small}} &
  CorpusRAN\_{\scriptsize Large} &
  \begin{tabular}[c]{@{}c@{}}7.91 \\ (std: 0.357)\end{tabular} &
  \begin{tabular}[c]{@{}c@{}}13.08\\ (std: 0.529)\end{tabular} &
  \begin{tabular}[c]{@{}c@{}}61.30 \\ (std: 3.114)\end{tabular} \\ \hline  
 &
  w/o fine-tuned &
  8.55 &
  12.30 &
  53.50 \\ \cline{2-5} 
 &
  \cellcolor[HTML]{EFEFEF}CorpusBAL\_{\scriptsize Large} &
  \cellcolor[HTML]{EFEFEF}\begin{tabular}[c]{@{}c@{}}3.24 \\ (std: 0.221)\end{tabular} &
  \cellcolor[HTML]{EFEFEF}\begin{tabular}[c]{@{}c@{}}7.89 \\ (std: 0.433)\end{tabular} &
  \cellcolor[HTML]{EFEFEF}\begin{tabular}[c]{@{}c@{}}\textbf{42.20} \\ (std: 1.483)\end{tabular} \\ \cline{2-5} 
\multirow{-3}{*}{CorpusRAN\_{\scriptsize Small}} &
  CorpusRAN\_{\scriptsize Large} &
  \begin{tabular}[c]{@{}c@{}}\textbf{3.02} \\ (std: 0.103)\end{tabular} &
  \begin{tabular}[c]{@{}c@{}}\textbf{7.41} \\ (std: 0.379)\end{tabular} &
  \begin{tabular}[c]{@{}c@{}}44.20 \\ (std: 1.483)\end{tabular} \\ \hline
\end{tabular}%
}
\end{table}

\begin{table}[htbp!]
\centering
\caption{T-test of the CorpusBAL and CorpusRAN ASR results}
\label{table:ASR_results_ttest}
\begin{tabular}{|c|ccc|}
\hline& \multicolumn{3}{c|}{p-value}\\ \hline
Testing data     & \multicolumn{1}{c|}{PER}     & \multicolumn{1}{c|}{CER}     & SER     \\ \hline
\rowcolor[HTML]{EFEFEF} CorpusBAL\_{\scriptsize Large} & \multicolumn{1}{c|}{0.00022} & \multicolumn{1}{c|}{0.00006} & 0.00617 \\ \hline
CorpusRAN\_{\scriptsize Large} & \multicolumn{1}{c|}{0.00051} & \multicolumn{1}{c|}{0.18967} & 0.03256 \\ \hline \hline
\rowcolor[HTML]{EFEFEF} CorpusBAL\_{\scriptsize Small} & \multicolumn{1}{c|}{0.00008} & \multicolumn{1}{c|}{0.00305} & 0.02148 \\ \hline
CorpusRAN\_{\scriptsize Small} & \multicolumn{1}{c|}{0.07949} & \multicolumn{1}{c|}{0.09961} & 0.06559 \\ \hline
\end{tabular}
\end{table}

Table~\ref{table:ASR_validation} compares the performance of best model selection using different validation sets. The ASR model was fine-tuned on CorpusBAL\_{\footnotesize Small} and tested on CorpusBAL\_{\footnotesize Large} and CorpusRAN\_{\footnotesize Large}. The best model was selected using a syllable-balanced set (cf. valid:bal in Table~\ref{table:ASR_validation}) or a randomly selected sentences set (cf. valid:ran in Table~\ref{table:ASR_validation}). The results show that the ASR model selected by a syllable-balanced validation set yields lower CER and SER than the ASR model selected by a randomly selected validation set. 

\begin{table}[htbp!]
\centering
\caption{ASR performance using different validation sets}
\label{table:ASR_validation}
\resizebox{\columnwidth}{!}{%
\begin{tabular}{|c|c|c|c|c|}
\hline
Testing data &
  Training data &
  PER &
  CER &
  SER \\ \hline
 &
  \cellcolor[HTML]{EFEFEF}\begin{tabular}[c]{@{}c@{}}CorpusBAL\_{\scriptsize Small} \\ (valid:bal)\end{tabular} &
  \cellcolor[HTML]{EFEFEF}\begin{tabular}[c]{@{}c@{}}9.658\\ (std: 0.259)\end{tabular} &
  \cellcolor[HTML]{EFEFEF}\begin{tabular}[c]{@{}c@{}}\textbf{15.544}\\ (std: 0.212)\end{tabular} &
  \cellcolor[HTML]{EFEFEF}\begin{tabular}[c]{@{}c@{}}\textbf{67.648}\\ (std: 0.957)\end{tabular} \\ \cline{2-5} 
\multirow{-2}{*}{CorpusBAL\_{\scriptsize Large}} &
  \begin{tabular}[c]{@{}c@{}}CorpusBAL\_{\scriptsize Small}\\ (valid:ran)\end{tabular} &
  \begin{tabular}[c]{@{}c@{}}\textbf{9.630}\\ (std: 0.263)\end{tabular} &
  \begin{tabular}[c]{@{}c@{}}15.622\\ (std: 0.274)\end{tabular} &
  \begin{tabular}[c]{@{}c@{}}67.898 \\ (std: 0.672)\end{tabular} \\ \hline
 &
  \cellcolor[HTML]{EFEFEF}\begin{tabular}[c]{@{}c@{}}CorpusBAL\_{\scriptsize Small}\\ (valid:bal)\end{tabular} &
  \cellcolor[HTML]{EFEFEF}\begin{tabular}[c]{@{}c@{}}4.885\\ (std: 0.062)\end{tabular} &
  \cellcolor[HTML]{EFEFEF}\begin{tabular}[c]{@{}c@{}}\textbf{9.244}\\ (std: 0.141)\end{tabular} &
  \cellcolor[HTML]{EFEFEF}\begin{tabular}[c]{@{}c@{}}\textbf{47.922}\\ (std: 0.518)\end{tabular} \\ \cline{2-5} 
\multirow{-2}{*}{CorpusRAN\_{\scriptsize Large}} &
  \begin{tabular}[c]{@{}c@{}}CorpusBAL\_{\scriptsize Small} \\ (valid:ran)\end{tabular} &
  \begin{tabular}[c]{@{}c@{}}\textbf{4.870}\\ (std: 0.132)\end{tabular} &
  \begin{tabular}[c]{@{}c@{}}9.250\\ (std: 0.168)\end{tabular} &
  \begin{tabular}[c]{@{}c@{}}47.976\\ (std: 0.445)\end{tabular} \\ \hline
\end{tabular}%
}
\end{table}

\section{Conclusion}
In this paper, we first present a statistical analysis of Mandarin Chinese acoustic units based on a large corpus of news texts collected from the internet. We then proposed the BASPRO system that selects sentences from a large text corpus to compose a syllable-balanced recording script with similar statistics. The experimental results showed that the BASPRO system can effectively produce a syllable-balanced script based on the designed fitness function. Using BASPRO, we obtained a recording script called \emph{TMNews}. Subsequently, we used TTS systems to convert sentences in the \emph{TMNews} script into utterances to form a speech corpus. Through SE and ASR experiments evaluated on speech corpora based on different recording scripts, we confirmed that SE and ASR models trained on a syllable-balanced speech corpus based on the \emph{TMNews} script outperformed those trained on a randomly formed speech corpus. In this study, we primarily focused on the design of audio-recording scripts rather than the audio recordings. There are too many variations in the recorded utterances, such as the recording device and the gender, age, and accent of the speaker. Therefore, the recording setting is beyond the scope of this study, and we used synthetic speech with relatively simple characteristics for the SE and ASR evaluation experiments. Furthermore, the data-processing phase does not ensure that every candidate sentence is logical and appropriate from a human perspective. Therefore, manual screening is required during the postprocessing phase. In the future, we hope to develop a method that better reflects human understanding of sentence semantics and reduces human involvement in corpus design.


\appendix 


\begin{table*}
\centering
  \label{table:pypinyin}
  \includegraphics[width=0.9\textwidth]{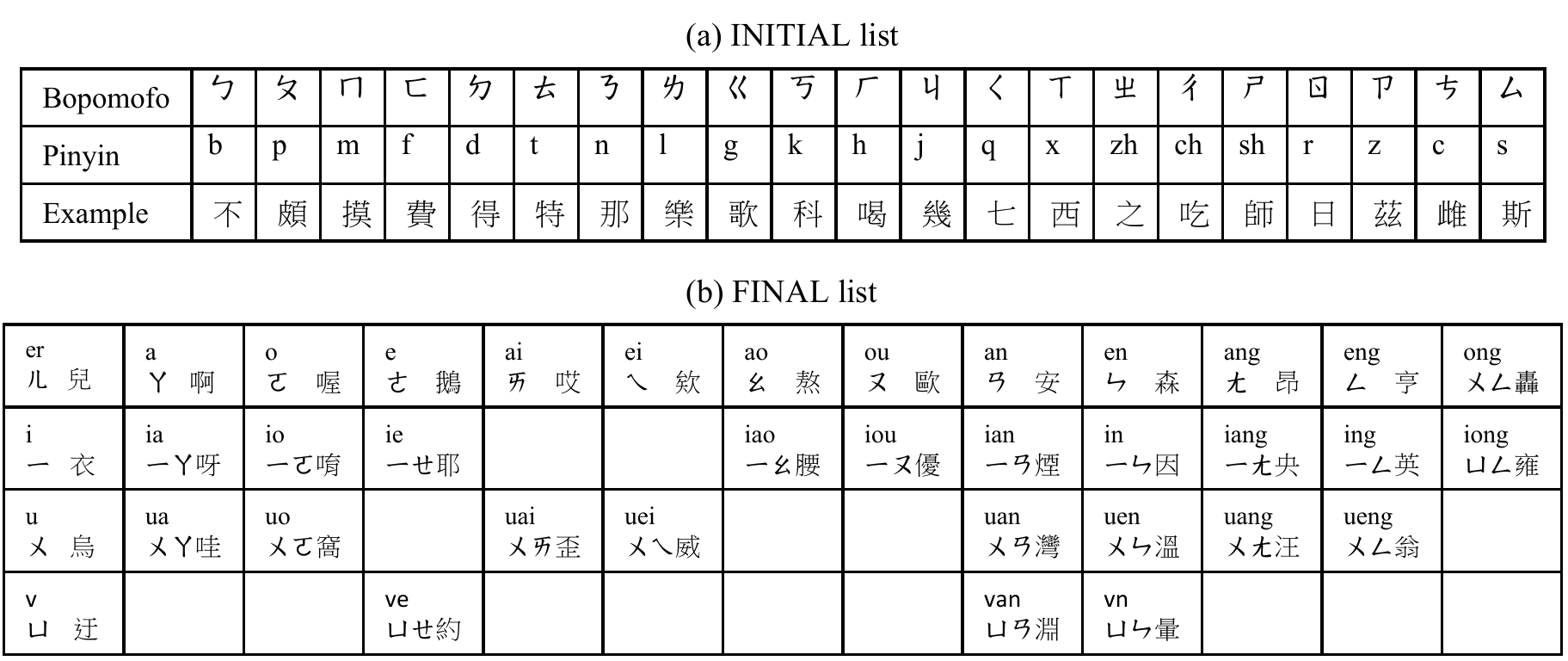}
   \caption{The INITIAL and FINAL list in the Pypinyin tool}
\end{table*}

\begin{figure*}
\centering 
\centerline{
\includegraphics[width=1.0\textwidth]{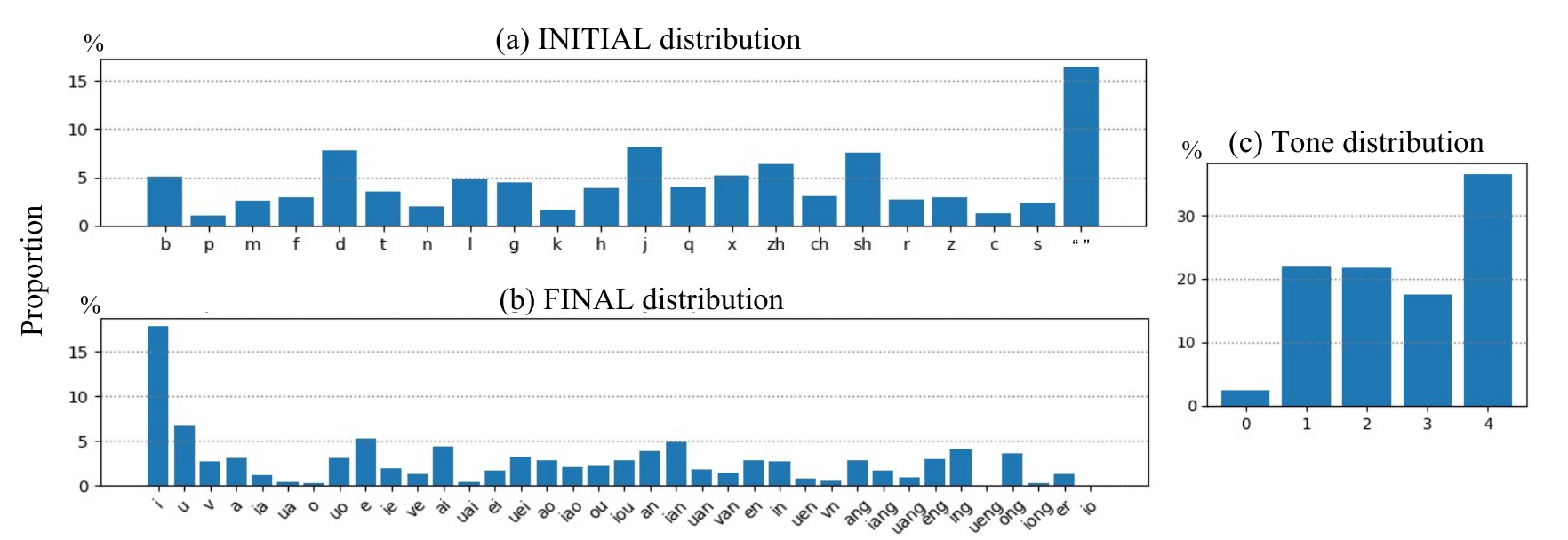}}
\caption{The INITIAL, FINAL, and tone distribution in news articles crawled from five major news media in Taiwan in 2021. `` '' in (a) refers to the syllable pronunciation without INITIAL.
 } 
\label{fig:real_world_dis}
\end{figure*}


\bibliographystyle{ieeetr}
\bibliography{reference}
\end{document}